*DISP '19, Oxford, United Kingdom*
*ISBN: 978-1-912532-09-4*# New Radon Transform Based Texture Features of Handwritten Document

Rustam Latypov
Kazan Federal University
Kazan, Russia
+79172808040
Roustam.Latypov@kpfu.ru

Evgeni Stolov
Kazan Federal University
Kazan, Russia
+78432387473
ystolov@kpfu.ru## ABSTRACT
**1.**     In this paper, we present some new features describing the handwritten document as a texture. These features are based on the Radon transform. All values can be obtained easily and suit for the coarse classification of documents.

## Keywords
Handwritten document, texture, Radon transform.## 2.     INTRODUCTION
Recently, digital documents are crowding out handwritten ones from everyday life. Even so, the problem of classifying a bundle of manuscripts meets often yet. The number of old manuscripts accumulated in various collections like libraries and archives is huge so developing new algorithms helping the scientists in dating, classification, and recognition of handwritten documents is still actual.

As a practical example, we give the following case. Let a teacher collects sheets with examination works of students, he/she must assemble the sheets belonging to the same person. The answers can take more than a single page and some students forget to designate its/her name on all pages. This is an example of the problem of determining the authorship of a manuscript. A more serious problem is to recognize a not subscribed manuscript or a small chunk of a manuscript containing text that was found in the museum [1]. The style and form of writing evolve over time and the way they have progressed enables researchers to identify the periods and the places in which a document was created. The common context of the all mentioned problems is a rather small size of the manuscript under consideration.

Currently, there are many papers dedicated to the automatic identification of the author of a handwritten document via templates of labeled examples [2-5]. As a rule, exact identification has to take into account various way of writing of particular letters. On the other hand, if there is a restricted number of versions some global features of the document, which may be thought of as a texture, can be exploited. Many authors use the Radon transform of a selected fragment of the document for obtaining a source data for further manipulation in the classification problem. For example, in [6] this idea is implemented as a software application intended for classification of signatures. On the base of the Radon transform of two signatures, a vector distance between two objects is calculated. The final decision is accepted after DWT transform of the obtained vector.

In our paper, we develop some new features describing the handwritten document as a texture. These features are received from a document using the Radon transform.

The paper is organized as follows. First, we give the introduction. Some definitions and ideas about manuscript features are in the next section. Then we describe the main theoretic as well as experimental results and make conclusions.

## 3.     GENERALIZED SLANT
Handwriting analysis considers different features concerning the motion and pressure of the hand, as well as the shape of the different symbols and the spatial relationship among them [7,8]. The slant of letters in the handwritten text is a substantial feature which is exploited for identification purposes. That value is related to a single letter or to a few connected letters in a line. Most of the existing methods for calculating angular measurements are based on the transformation of the grapheme, leading to its standard form [9]. In [10], a new procedure for finding the slant was proposed; we describe the idea of the method.

The source document has a black-white format and is present by a $2D$ array $M$, where all black pixels equals 1 and all white pixels equals 0. The $2D$ array $L$ in Fig. 1 has values 1 at the line that is rotated at the angle $t$ relative to the X-axis. While correlation counting between arrays $M$ and $L$ one obtains the result of Radon transform of the document as a projection along the direction with the angle $t$. All the technical details providing an effective procedure are presented in the paper [10].

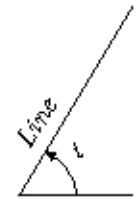

**Figure 1. The straight line at an angle *t* relative to the *X*-axis.**

If the array $M$ contains just a line of the document or a part of a line the slant in the line can be found by means of the following procedure. The result of the correlation between $M$ and $L$ with a given '$t$' is a function $f(x,t)$, where '$x$' defines shift of $L$ relative to beginning $M$. Let us normalize the value $f(x,t)$ dividing it to the value $S(t) = \sum_x f(x,t)$ and keeping the previous designation for $f(x,t)$. Then calculate the entropy

$$Entr(t) = -\sum_x f(x,t) \cdot \log(f(x,t)).$$

It can be proved [10] that $S(t)$ is independent of $t$, whereas the distribution of values of $f(x,t)$ depends on $t$. If there is $t^*$ close to the real slant of letters in the line then the values of $f(x,t^*)$ are

DOI: http://dx.doi.org/10.17501........................................

concentrated in a restricted set of arguments and $Entr(t^*)$ will be a local minimum of $Entr(t)$ [11].

In praxis, it may be troublesome to select automatically a single line in a handwritten document if the lines are not horizontal. Although there are a lot of papers relating to the problem of selecting a line from the text all the known solutions need for extra resources. Instead, we suggest calculating a value that we call "generalized slant" in a strip of the text for investigation. To this end, we calculate a correlation between the whole strip and the line $L$ for all $t$ chosen with some step. Then we find the local minimum of $Entr(t)$. A technical restriction on the implementation of the procedure is as follows: the width of the array $M$ must be at least 5 times more than its height [9]. To fit this restriction one can divide the source strip into strips of the same height and calculate the correlation between $L$ and small strips stretched in a row. The found position of the local minimum and the curve $Entr(t)$ itself can be viewed as a feature of the strip.

## 4. RADON TRANSFORM AND CORRELATION

It was mentioned above that Radon transform can be a source for the following processing procedure. Here we present a new example of such implementation. As before, we suppose that there is a strip from a handwritten document. A global feature of the strip must be found that can be used for further classification. An additional restriction on the procedure is as follows: it must not require many resources. Consider the following Algorithm 1.

---

READ *Step*, *M*

READ *Row*, *Col*  height and width  of *M*

OBTAIN *K*= *[Row/Step]*  integral part

OBTAIN *K* small arrays dividing *M* into strips of height *Step*

OBTAIN array *Mjoin* joining  *K* arrays in one row

SET *Threshold = Step/2*

SET empty *Sequence*

FOR  EACH  *Column*  IN  *Mjoin*

   CALCULATE *Val* = SUM(*Column*)

   IF *Val < Threshold*

      *Sequence* ADD 0

   ELSE

  *Sequence* ADD 1

   ENDIF

ENDFOR

SHOW autocorrelation function *Auto* of *Sequence*

---

**Algorithm 1. Finding autocorrelation function.**

Here the strip under investigation is presented as a black-white 2D array $M$ where 1's relate to the black pixels in the text. The array $M$ is divided into small $K$ arrays of the height $Step$. The algorithm produces a $Sequence$ that consists of 0's and 1's . $Sequence[x] = 0$ if the sum of ones in $Column$ with number $x$ in $Mjoin$ contains fewer ones than half of the height $Step$. Otherwise, $Sequence[x] = 1$. Calculation of the number of ones in a column can be viewed as a special case of Radon transform. If the strip is a single line of the text, the function *Auto* shows a regularity in the distribution of spaces between words and letters. In the case considered in the paper, the *Auto* depicts a kind of integral feature combining distribution all spaces. It will be shown that the plot of *Auto* substantially depends on the parameter *Step*. We demonstrate all the methods presented in the paper, using as an example the processing of two strips taken from examination sheets filled in by students during an exam in Math.

## 5. EXPERIMENTAL RESULTS

Two strips from examination papers of different students were selected for the experiment. They are presented in Fig. 2 and Fig. 3.

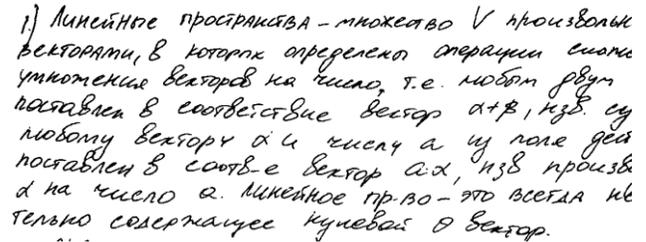

**Figure 2. The first piece of handwritten text under experiment.**

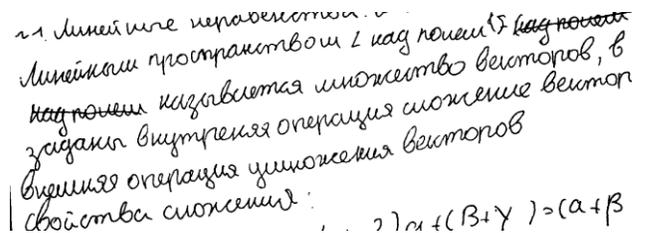

**Figure 3. The second piece of handwritten text under experiment.**

### 5.1 Calculation of the generalized slant

Both the strips are divided into small strips of height 30 pixels and 50 pixels. The entropy curves are plotted in Fig. 4 and Fig. 5.

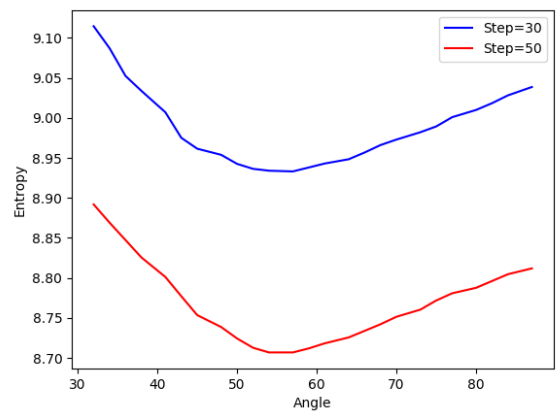

**Figure 4. Entropy for the first piece of handwritten text.**

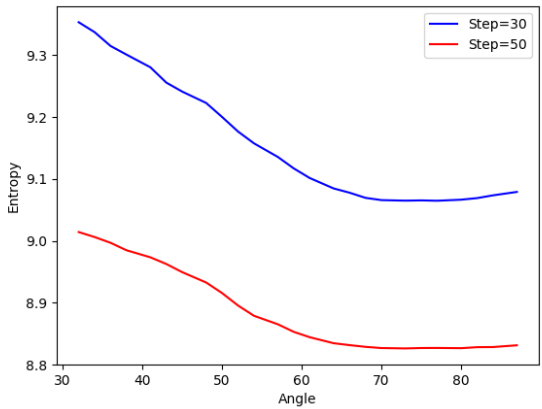

**Figure 5. Entropy for the second piece of handwritten text.**

Two curves (blue line - 30 pixels, red line - 50 pixels) have almost the same shape and the same local minimum. The minimum of the entropy for the first picture is close to 57 degrees and for the other is close to 74 degrees.

## 5.2 Correlation of the Radon transform

Both the chosen strips were processed by the Algorithm 1. Fig. 6 and Fig. 7 present values of the autocorrelation functions calculated for *Step* value equal to 15.

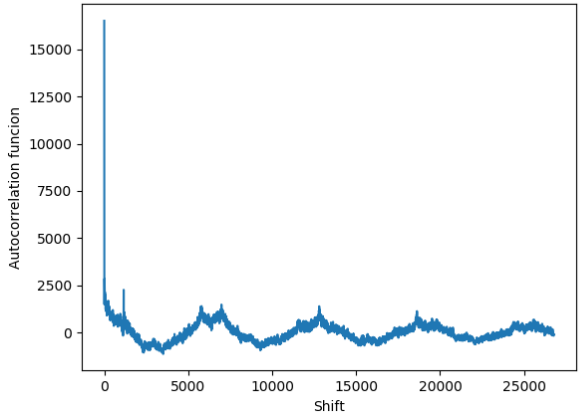

**Figure 6. Autocorrelation for the first piece of handwritten text.** *Step=15*.

The former plot relates to the strip in Fig. 2 and the latter belongs to one in Fig. 3. The main period for the plot in Fig. 6 points to the length of line in the strip and is proportional to it since the lines are almost horizontal. In Fig. 7 period depicts autocorrelation of positions of spaces in the text.

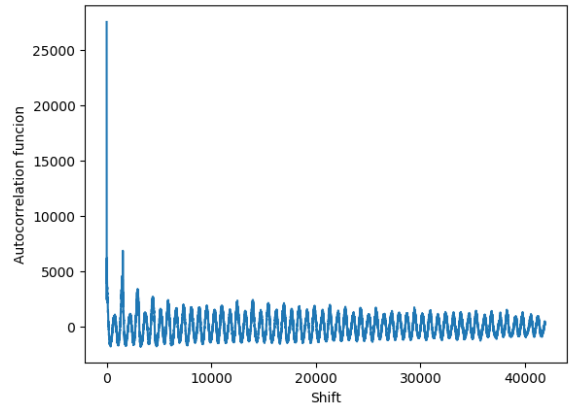

**Figure 7. Autocorrelation for the second piece of handwritten text.** *Step=15*.

Fig. 8 and Fig. 9 present values of the autocorrelation functions for two chosen pictures calculated for different values of *Step* in compressed form. The distinctions between these graphs are evident and they can be leveraged for classification. A method of estimating a digital distance between such the graphs is a subject for the further research of the authors.

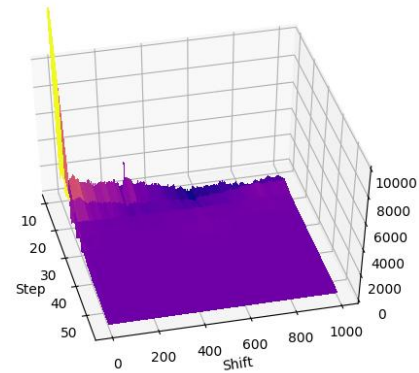

**Figure 8. Autocorrelation for the first piece of handwritten text.**

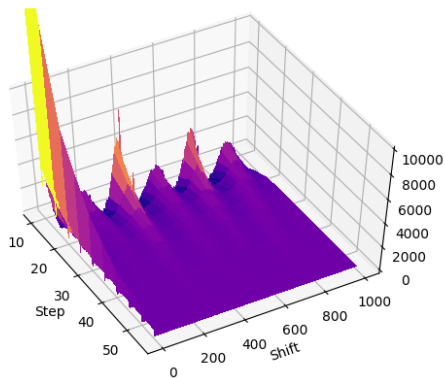

**Figure 9. Autocorrelation for the second piece of handwritten text.**

## 6. CONCLUSIONS

In this paper, we have proposed two original methods for effective calculation of new texture parameters of fragments of handwritten papers. The methods assumed to be implemented for the initial classification of manuscripts.

## 7. ACKNOWLEDGMENTS

The work is performed according to the Russian Government Program of Competitive Growth of Kazan Federal University. This work is also supported by the RFBR grant №18-47-16005.